%% file: main.tex
\RequirePackage[svgnames]{xcolor}

\documentclass[11pt,letterpaper]{mystyle}
\usepackage{bxcoloremoji}

\usepackage[all]{hypcap}
\usepackage[svgnames]{xcolor}
\usepackage[comma,authoryear,compress]{natbib}
\usepackage{hyperref}[citecolor=lightblue]

\hypersetup{
    colorlinks = true,
    citecolor = {YaleBlue},
}

\usepackage{algorithm}
\usepackage{algorithmicx}
\usepackage{algpseudocode}
\usepackage{microtype}
\usepackage{graphicx}
\expandafter\def\csname ver@subfig.sty\endcsname{}
\usepackage{booktabs} %
\usepackage{float}
\usepackage{bigstrut}
\usepackage{multirow}
\usepackage{amsmath}
\usepackage{amssymb}
\usepackage{mathtools}
\usepackage{amsthm}
\usepackage{mathrsfs}
\usepackage{nicefrac}
\usepackage{dsfont}
\usepackage{enumitem}
\usepackage{subcaption}
\usepackage{graphicx,subfig}
\usepackage{cleveref}
\usepackage{bxcoloremoji}
\usepackage{float}
\usepackage{fontawesome5}

\usepackage[utf8]{inputenc} %
\usepackage[T1]{fontenc}    %
\usepackage{hyperref}       %
\usepackage{url}            %
\usepackage{booktabs}       %
\usepackage{amsfonts}       %
\usepackage{nicefrac}       %
\usepackage{microtype}      %
\usepackage{graphicx}
\usepackage{subcaption} 
\usepackage{amssymb}
\usepackage{fdsymbol}
\usepackage{wrapfig}
\usepackage{lipsum}
\usepackage{enumitem}
\usepackage{stackengine}
\usepackage[font=small,labelfont=bf]{caption}
\usepackage{color}
\usepackage{adjustbox}

\usepackage{rotating}
\usepackage{makecell}

\input{macro}

\newtcolorbox{AIbox}[2][]{aibox,title=#2,#1}
\definecolor{lightblue}{rgb}{0.22,0.45,0.70}%
\definecolor{Gray}{gray}{0.95}
\definecolor{Cornsilk}{rgb}{1.0, 0.97, 0.86}

\usepackage{amsmath}

\usepackage[all]{hypcap}

\graphicspath{{figures/}}

\definecolor{groupgray}{RGB}{245,245,245}
\definecolor{highlightblue}{RGB}{215,242,248}
\newcommand{\pos}[1]{\textcolor{green!60!black}{\scriptsize #1}}
\newcommand{\negg}[1]{\textcolor{red}{\scriptsize #1}}

\title{Mixture-of-Memories: Low-Overhead Language Modeling with Modular Knowledge Injection}

\lab{GenMI Lab, MBZUAI}
\reportdate{\today}

\runningtitle{Mixture-of-Memories: Low-Overhead Language Modeling with Modular Knowledge Injection}

\author{
Yifan Lu$^{1}$, Rubin Wei$^{2}$, Ziyun Zou$^{1}$, Junhong Liang$^{1}$, Haolin Yang$^{1}$, Xinlin Zhuang$^{3}$, Abhijit Das$^{1}$, Adinath Madhavrao Dukre$^{1}$, Shadab Khan$^{4}$, Zhouhan Lin$^{2}$, Imran Razzak$^{1}$
}

\affil[1]{MBZUAI, Abu Dhabi, UAE}
\affil[2]{LUMIA Lab, School of Artificial Intelligence, Shanghai Jiao Tong University, China}
\affil[3]{The Chinese University of Hong Kong, Hong Kong}
\affil[4]{ADIA Lab, Abu Dhabi, UAE}

\correspondingauthor{Imran Razzak; Email: \href{mailto:Imran.Razzak@mbzuai.ac.ae}{Imran.Razzak@mbzuai.ac.ae}}

\begin{document}

\input{sections/abstract}

\maketitle

\par
\newsavebox{\teasercapbox}
\sbox{\teasercapbox}{\begin{minipage}{\textwidth}
\captionsetup{skip=0pt}

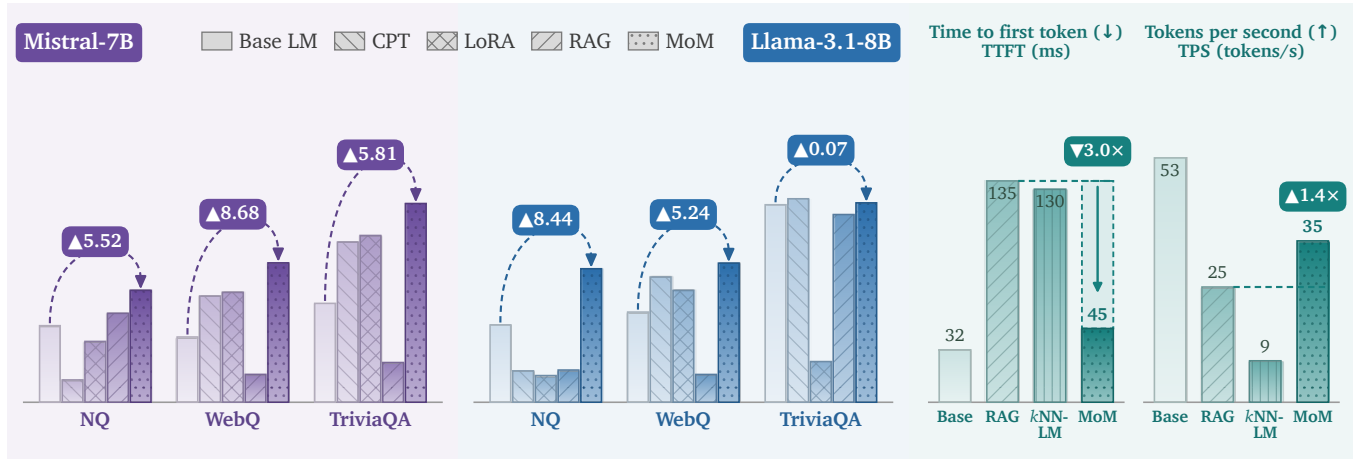
\captionof{figure}{Performance and efficiency comparison. \textbf{Left and middle}: accuracy on three QA benchmarks with Mistral-7B and Llama-3.1-8B; bars are scaled within each benchmark. MoM improves over the base LM on every benchmark, whereas parametric methods (CPT, LoRA) and non-parametric retrieval (RAG) each lower at least one. \textbf{Right}: inference efficiency on a 7B backbone, measured by time to first token (TTFT, $\downarrow$ lower is better) and tokens per second (TPS, $\uparrow$ higher is better). Relative to RAG and $k$NN-LM, MoM lowers TTFT by 2.9--3.0$\times$ and raises TPS by 1.4--3.9$\times$.}
\label{fig:teaser}
\end{minipage}}
\newlength{\teaserheight}
\setlength{\teaserheight}{\dimexpr\pagegoal-\pagetotal-\ht\teasercapbox-\dp\teasercapbox-\parskip-16pt\relax}
\noindent\makebox[\textwidth]{\includegraphics[width=\textwidth,height=\teaserheight,keepaspectratio]{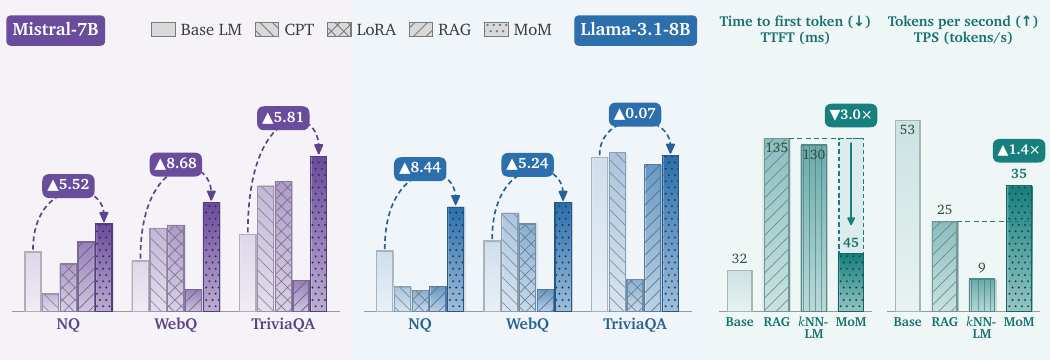}}\par\nopagebreak
\vspace{4pt}\nopagebreak
\noindent\usebox{\teasercapbox}

\newpage

\input{sections/content}
\clearpage
\bibliography{main}

\clearpage
\appendix
\input{sections/appendix}

\end{document}

%% file: macro.tex
\definecolor{blanchedalmond}{rgb}{1.0, 0.92, 0.8}
\definecolor{carmine}{rgb}{0.59, 0.0, 0.09}
\definecolor{lightblue}{rgb}{0.22,0.45,0.70}%

\renewcommand{\mathbf}{\boldsymbol}

\makeatletter
\def\Ddots{\mathinner{\mkern1mu\raise\p@
\vbox{\kern7\p@\hbox{.}}\mkern2mu
\raise4\p@\hbox{.}\mkern2mu\raise7\p@\hbox{.}\mkern1mu}}
\makeatother

\definecolor{amaranth}{rgb}{0.9, 0.17, 0.31}
\definecolor{antiquebrass}{rgb}{0.8, 0.58, 0.46}
\definecolor{antiquefuchsia}{rgb}{0.57, 0.36, 0.51}
\definecolor{chromeyellow}{rgb}{0.31, 0.47, 0.26}



%% file: sections/abstract.tex
\begin{abstract}
Large language models store substantial factual knowledge in their parameters, yet this knowledge is difficult to control, update, and reliably elicit at inference time. Existing memory-augmented approaches typically rely on a single auxiliary mechanism, such as retrieval, parameter updates, or neural memory, and therefore apply the same form of correction to tokens with very different knowledge demands. We propose Mixture-of-Memories (MoM), a memory-augmented language modeling framework that separates routed neural memory from writable factual memory. MoM combines a frozen base LM with two neural memory experts: a latent memory that provides low-latency hidden-state corrections and a lexical memory that provides a token-space knowledge pathway. A lightweight Budget Router sets per-token mixture weights from uncertainty and disagreement features, replacing the fixed interpolation of prior memory modules. For post-hoc updates, an external explicit memory stays outside the router, because a router trained before facts are written learns to down-weight it; new facts are appended without retraining any component and used only when an answer-level gate accepts the retrieved fact. Experiments across language modeling, factual QA, update analysis, and efficiency show that MoM improves both predictive accuracy and updateability while retaining low inference overhead.

\vspace{5mm}


\coloremojicode{1F4E7} \textbf{Contact}: \href{mailto:Yifan.Lu@mbzuai.ac.ae}{Yifan.Lu@mbzuai.ac.ae}

\end{abstract}

%% file: sections/content.tex
\section{Introduction}
\label{sec:intro}

Large language models (LLMs) encode a substantial amount of world knowledge in their parameters~\citep{allen2025physics}. This capability follows from large-scale next-token prediction and is one reason pretrained LMs generalize well. Parametric knowledge, however, is a poor interface for knowledge-intensive prediction. Parameters are optimized to capture broad statistical regularities, not to give explicit control over how individual facts are stored, accessed, or updated. The mechanism that handles common linguistic and factual patterns therefore becomes unreliable when a prediction depends on precise, infrequent, or newly introduced knowledge~\citep{chang2024acquire,gekhman2024finetuning,ovadia2024finetuning}, which leaves a gap between what an LLM absorbs during pretraining and what it can reliably use at inference time.

This gap has at least three sources. A fact may be weakly represented because it is under-supported in pretraining, difficult to recover because the context does not elicit it, or stale because it changed after training. Existing approaches address these failure modes only partially. Parameter-side methods internalize new information into model weights, but frequent or repeated updates remain difficult to support cleanly~\citep{selfparam}. External retrieval and token-level memory make updates easier but add search and context-processing overhead at inference time~\citep{cao2026memory}. Latent-space memory compresses information into hidden representations and can improve retention, but it is tied to model-specific representations~\citep{memoryllm,mplus}. Plug-and-play parametric memories imitate non-parametric retrievers with lightweight auxiliary modules, which improves portability and inference efficiency, yet they still expose a single auxiliary path with fixed interpolation~\citep{cao2026memory,mlpmemory}.

Because all of these mechanisms ultimately act through next-token prediction at inference time, their differences matter only insofar as they change the local prediction problem during generation. The need for extra knowledge is not uniform across positions. At many steps, the base LM is already sufficient: the next token is governed mainly by local syntax, common lexical patterns, or highly reinforced knowledge, so auxiliary support brings little benefit. The challenging cases are concentrated on a much smaller subset of positions where prediction depends on precise factual content, such as entities, numbers, dates, and other knowledge-bearing slots~\citep{yao2025seakr}. Even there, the relevant deficiency is not always the same. Some errors reflect weak parametric support and call for a lightweight correction to the base prediction; others reflect a need for more explicit token-level guidance or direct access to facts introduced after pretraining. This makes a single global augmentation rule difficult to justify. If the same auxiliary correction is applied uniformly throughout generation, the model is forced to treat easy and difficult positions alike, and to use the same form of support for qualitatively different failure modes. A more natural formulation is to let knowledge support vary with the local prediction state, invoking different auxiliary paths only where they are likely to reduce token-level error.

To bridge the gap, we propose Mixture-of-Memories (MoM), a memory-augmented language modeling framework that separates memory for stable prediction from memory for post-hoc updates. MoM combines a frozen base LM with two routed neural memory experts: a latent memory (LatM) that provides low-latency, retrieval-like corrections in hidden space, and a lexical memory (LexM) that operates in token space and is therefore less coupled to the backbone's representations. A lightweight Budget Router assigns token-wise mixture weights over these stable sources from uncertainty and disagreement features. Mutable knowledge goes to an explicit memory (ExpM), an external store that is kept outside the router; it retrieves a stored fact once per question and injects its answer only when an answer-level gate judges the fact reliable. Facts are added by appending entries, without retraining the backbone, the neural memories, or the router.

Our contributions are as follows:
\begin{itemize}
\item We propose Mixture-of-Memories, which composes a frozen base language model with a latent and a lexical memory through a lightweight token-wise Budget Router.
\item We keep mutable knowledge in an explicit memory outside the router, so new facts can be written without retraining the backbone, the neural memories, or the router, and take effect immediately.
\item Extensive experiments show that MoM reduces WikiText-103 perplexity on GPT2-large from 10.45 to 9.37, improves the average factual QA score over frozen backbones by 15.6\% on Mistral-7B and 10.9\% on Llama-3.1-8B, and lowers time to first token by 2.9--3.0$\times$ relative to retrieval-augmented baselines on a 7B backbone.
\end{itemize}

\section{Related Work}
\label{sec:related_work}

\paragraph{Retrieval-augmented generation.}
Retrieval-augmented generation grounds predictions in external knowledge. RAG~\citep{lewis2020rag} retrieves documents or passages, RETRO~\citep{borgeaud2022retro} retrieves corpus chunks, and $k$NN-LM~\citep{DBLP:conf/iclr/KhandelwalLJZL20} retrieves nearest-neighbor continuations from a datastore of activations. Recent work post-trains the reader to make better use of retrieved evidence, through hierarchical-thought instruction tuning in HIRAG~\citep{jiao2025hirag} or reinforcement learning with a curriculum in RAG-RL~\citep{huang2025ragrl}. These methods keep knowledge explicit and easy to refresh, but they shift cost to inference: online search adds latency, retrieved evidence lengthens the context, and large indexes or datastores require storage. MoM amortizes retrieval-like corrections into neural memory experts that are invoked through token-wise routing, and it reserves explicit lookup for a small store of written facts.

\paragraph{Parametric long-term memory.}
Parametric memory methods reduce the serving cost of retrieval by compressing corpus-level knowledge into learned auxiliary modules. MLP Memory~\citep{mlpmemory} uses an external MLP to mimic non-parametric retrieval, and Memory Decoder~\citep{cao2026memory} attaches a plug-and-play decoder memory to a frozen language model. Both make memory access fast and model-compatible, but compressing knowledge into parameters gives up direct writability, one of the main advantages of retrieval: adding or revising facts requires retraining, re-distillation, or rebuilding the memory. MoM keeps parametric memories for stable, low-latency modeling and delegates factual updates to a separate explicit memory.

\paragraph{Conditional computation and knowledge editing.}
Sparse mixture-of-experts models route tokens to experts to scale capacity~\citep{DBLP:conf/iclr/ShazeerMMDLHD17,fedus2022switch,du2022glam}. MoM instead routes among heterogeneous knowledge sources, namely the base LM, a hidden-space memory, and a token-space memory, for conditional memory access. For editing, MEND~\citep{DBLP:conf/iclr/MitchellLBFM22}, ROME~\citep{meng2022rome}, and MEMIT~\citep{DBLP:conf/iclr/MengSABB23} modify model weights, whereas SERAC~\citep{mitchell2022serac}, GRACE~\citep{hartvigsen2023grace}, and WISE~\citep{wang2024wise} store edits in auxiliary modules. ExpM is closest to the second group, but it writes facts as raw key--value entries without per-edit optimization.

\section{Methodology}
\label{sec:methodology}

\begin{figure}[t]
  \centering
  \includegraphics[width=\textwidth]{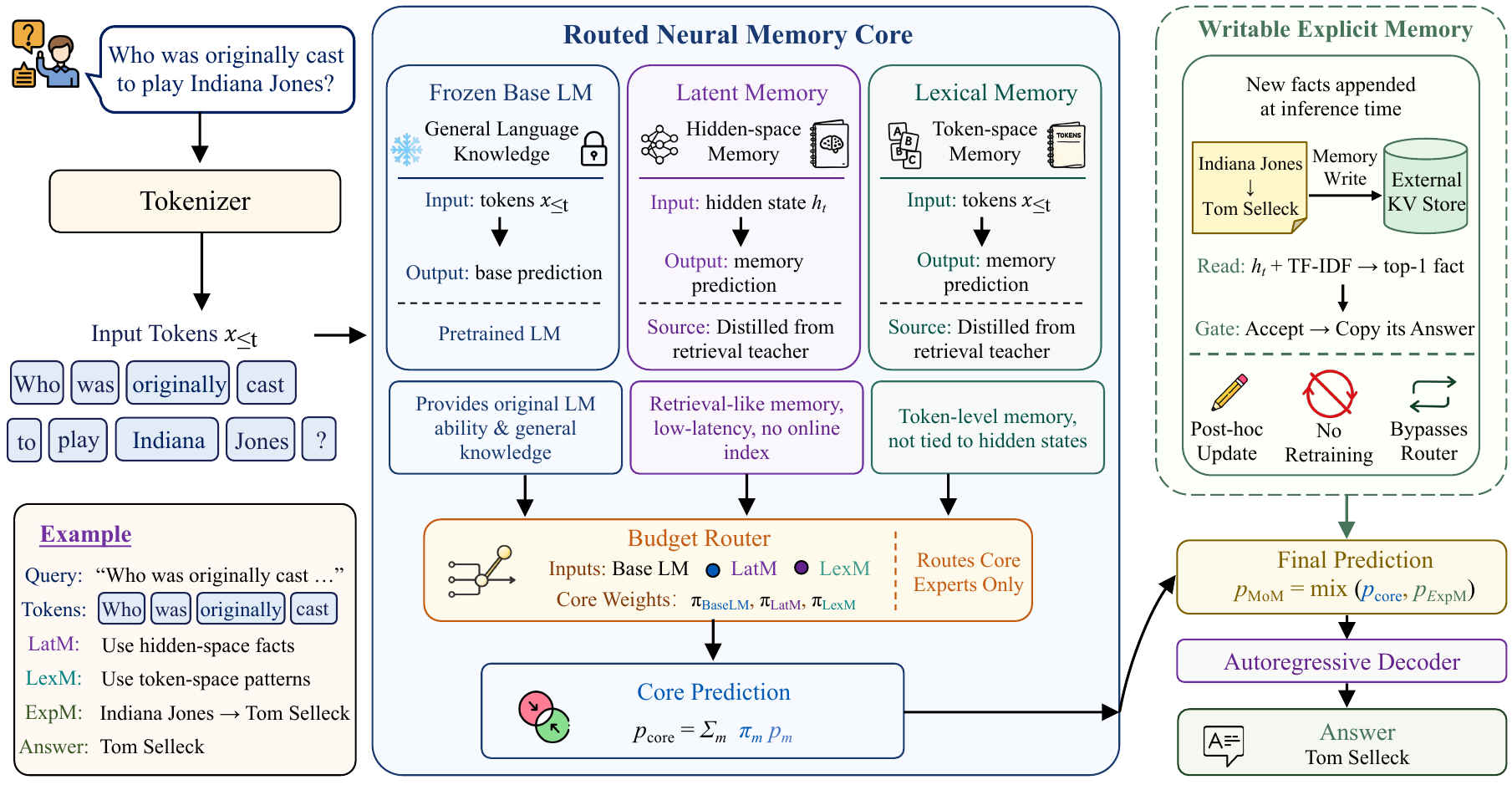}
  \caption{Mixture-of-Memories architecture. The input is tokenized and each next-token prediction is formed from three core sources: a frozen Base LM, a hidden-space Latent Memory (LatM), and a token-space Lexical Memory (LexM). The Budget Router produces per-token mixture weights over these sources. The writable Explicit Memory (ExpM) stores appended facts in an external key--value store and bypasses the router. For each question, it retrieves the fact and injects its answer only when an answer-level gate accepts it; otherwise the routed core predicts alone. New facts thus take effect without retraining.}
  \label{fig:framework}
\end{figure}

MoM is built on a separation principle: stable neural memories are routed, while mutable explicit knowledge is injected externally. The two kinds of memory serve different purposes. Neural memories are trained once and reused across inputs, which makes them suitable for token-level routing. Explicit memory is sparse, user- or time-dependent, and may be empty during router training but populated at test time. Placing such a store in the same router softmax lets the router learn that explicit memory is useless whenever the store is empty (Section~\ref{subsec:ablation_studies}). MoM therefore consists of a routed neural memory core and an external explicit memory overlay (Figure~\ref{fig:framework}).

Given a context $x_{\leq t}=(x_1,\ldots,x_t)$, the model predicts the next token $y_t=x_{t+1}$. The routed core consists of three stable sources,
\begin{equation}
\mathcal{C}=\{\mathrm{Base},\mathrm{LatM},\mathrm{LexM}\},
\label{eq:mom_core_set}
\end{equation}
where $\mathrm{Base}$ is the frozen parametric memory of the language model, $\mathrm{LatM}$ is a latent-space memory distilled from retrieval behavior, and $\mathrm{LexM}$ is a token-space lexical memory.

\subsection{Design Overview}
\label{subsec:mom_overview}

MoM first builds a core distribution by routing among the stable sources:
\begin{equation}
p_{\mathrm{core}}(y_t \mid x_{\leq t})
=\sum_{m\in\mathcal{C}}\pi_m(t)\,p_m(y_t \mid x_{\leq t}),
\label{eq:mom_core_mixture}
\end{equation}
where $p_m$ is the distribution produced by source $m$ and $\pi_m(t)$ is a token-dependent routing weight. The conditioning of $p_m$ is module-specific: Base and LexM consume the token sequence, while LatM consumes a hidden representation extracted from the frozen base model.

The final prediction overlays the explicit memory:
\begin{equation}
p_{\mathrm{MoM}}(y_t \mid x_{\leq t})
=(1-\alpha_t)\,p_{\mathrm{core}}(y_t \mid x_{\leq t})
+\alpha_t\,p_{\mathrm{ExpM}}(y_t \mid x_{\leq t},\Delta),
\label{eq:mom_final_distribution}
\end{equation}
where $\Delta$ is the external key--value store. The coefficient $\alpha_t=d_t\lambda_t$ separates two decisions: $d_t\in\{0,1\}$ indicates whether ExpM's gate accepts the retrieved fact (Eq.~\ref{eq:mom_gate}), and $\lambda_t\in[0,1]$ sets the injection strength. When the gate rejects, $d_t=0$ and Eq.~\ref{eq:mom_final_distribution} reduces exactly to the routed core. Neither $d_t$ nor $\lambda_t$ is produced by the Budget Router.

\subsection{Routed Neural Memory Core}
\label{subsec:routed_core}

\paragraph{Base memory.}
The base source is a frozen autoregressive language model with parameters $\theta_{\mathrm{Base}}$. It produces logits $z_{\mathrm{Base}}(t)=F_{\theta_{\mathrm{Base}}}(x_{\leq t})$ and the next-token distribution
\begin{equation}
p_{\mathrm{Base}}(y_t\mid x_{\leq t})=\mathrm{softmax}\big(z_{\mathrm{Base}}(t)\big)_{y_t}.
\label{eq:mom_base_distribution}
\end{equation}
The base model also provides a hidden representation
\begin{equation}
h_t=\phi_{\theta_{\mathrm{Base}}}^{(\ell)}(x_{\leq t})
\label{eq:mom_hidden_state}
\end{equation}
from a fixed layer $\ell$. LatM and the ExpM encoders reuse hidden states from the same backbone forward pass, while LexM takes the token sequence as input.

\paragraph{Latent memory.}
LatM removes the online nearest-neighbor search of retrieval-augmented LMs. Instead of searching a datastore at every token, it learns a parametric approximation of retrieval-style corrections in the hidden space of the base model:
\begin{equation}
p_{\mathrm{LatM}}(y_t\mid h_t)=\mathrm{softmax}\big(G_{\mathrm{LatM}}(h_t)\big)_{y_t}.
\label{eq:mom_latm_distribution}
\end{equation}
$G_{\mathrm{LatM}}$ is a lightweight MLP-style memory module. It is trained by distilling retrieval-augmented teacher distributions and frozen afterwards, and it provides low-latency corrections when the base representation already contains enough signal to predict the needed retrieval behavior.

\paragraph{Lexical memory.}
LexM complements LatM by operating in token space rather than in the hidden space of the base model:
\begin{equation}
p_{\mathrm{LexM}}(y_t\mid x_{\leq t})=\mathrm{softmax}\big(\mathrm{Dec}_{\mathrm{LexM}}(x_{\leq t})\big)_{y_t}.
\label{eq:mom_lexm_distribution}
\end{equation}
Because LexM consumes token IDs rather than hidden states, it does not depend on the representation geometry of a particular base model. LatM is efficient but base-specific, whereas LexM can be attached to another backbone with a compatible vocabulary (Appendix~\ref{app:lexm_transfer}). After training, LexM is frozen and becomes a stable source of the routed core.

\subsection{Budget Router}
\label{subsec:mom_budget_router}

The Budget Router decides when each stable source should influence the next-token distribution. It outputs a distribution over the routed core:
\begin{equation}
\log \boldsymbol{\pi}(t)=\mathrm{LogSoftmax}\big(R_{\psi}(f_t)\big),
\label{eq:mom_router}
\end{equation}
where $\boldsymbol{\pi}(t)=[\pi_{\mathrm{Base}}(t),\pi_{\mathrm{LatM}}(t),\pi_{\mathrm{LexM}}(t)]$, $R_{\psi}$ is an MLP, and $f_t$ contains inexpensive uncertainty and disagreement features:
\begin{equation}
f_t=\big[\,H_{\mathrm{Base}}(t),\ \mathrm{margin}_{\mathrm{Base}}(t),\
\operatorname{JSD}(p_{\mathrm{Base}},p_{\mathrm{LatM}})(t),\
\operatorname{JSD}(p_{\mathrm{Base}},p_{\mathrm{LexM}})(t)\,\big].
\label{eq:mom_router_features}
\end{equation}
$H_{\mathrm{Base}}$ is the entropy of the base distribution, $\mathrm{margin}_{\mathrm{Base}}$ is the gap between its two largest token probabilities, and $\operatorname{JSD}$ measures disagreement between distributions. The core mixture is computed in log space with the log-sum-exp operator, $\operatorname{LSE}_{m}(a_m)=\log\sum_{m}\exp(a_m)$:
\begin{equation}
\log p_{\mathrm{core}}(y_t\mid x_{\leq t})=\operatorname{LSE}_{m\in\mathcal{C}}
\big[\log \pi_m(t)+\log p_m(y_t\mid x_{\leq t})\big].
\label{eq:mom_log_core}
\end{equation}
ExpM is deliberately absent from Eq.~\ref{eq:mom_router} because it is mutable and may be empty while the router is trained, so routing it would teach the router to suppress explicit memory.

\subsection{External Explicit Memory}
\label{subsec:mom_expm}

ExpM stores facts as writable entries. A fact consists of a question or prompt $x_i$ and its answer token sequence $a_i=(a_{i,1},\ldots,a_{i,M_i})$. Each entry keeps a dense key $k_i$, a TF-IDF vector $\tilde{k}_i$ of $x_i$, the answer $a_i$, and a source record. Adding knowledge appends entries to the store $\Delta$ and changes no parameter of the base model, LatM, LexM, the router, or the ExpM encoders.

\paragraph{Query and key encoders.}
ExpM reads the frozen backbone rather than raw text. At position $t$ it forms the feature
\begin{equation}
z_t=\big[\,h_t\,;\ \bar{h}_t\,\big],
\label{eq:mom_expm_feature}
\end{equation}
where $h_t$ is the backbone hidden state of Eq.~\ref{eq:mom_hidden_state} and $\bar{h}_t$ is the mean hidden state of the most recent valid tokens. Two separate towers $E_q$ and $E_k$ map this feature to L2-normalized vectors. The query $q=E_q(z_T)$ is taken at the last position $T$ of the question, where the answer starts, and the key of a fact, $k_i=E_k(z(x_i))$, is taken at the last position of its own question prefix.

\paragraph{Retrieval.}
ExpM retrieves once per question, before the answer is generated. Each entry is scored by a convex combination of dense and lexical similarity,
\begin{equation}
s_i=(1-\beta)\,\frac{q^{\top}k_i+1}{2}+\beta\,\cos\big(\tilde{q},\tilde{k}_i\big),
\label{eq:mom_expm_score}
\end{equation}
where $\tilde{q}$ is the TF-IDF vector of the question and $\beta\in[0,1]$ weights the lexical term. Rescaling the dense cosine to $[0,1]$ puts both terms on the same range. ExpM takes the highest-scoring entry $i^{\star}$ over the whole store as its answer candidate. The top-$k$ entries $\mathcal{N}$ under the same score serve to compute gate features.

\paragraph{Answer-level gate.}
Retrieval always returns a candidate, including for questions whose fact is not in the store, so ExpM must decide whether to trust it. A logistic-regression gate estimates the probability that the retrieved answer is correct:
\begin{equation}
g(x)=\sigma\big(\mathbf{w}^{\top}\boldsymbol{\phi}(x)+b\big),\qquad d=\mathbf{1}\big[g(x)\geq\eta\big].
\label{eq:mom_gate}
\end{equation}
The features $\boldsymbol{\phi}(x)$ are the retrieval score $s_{i^{\star}}(x)$, its margin over the second-best candidate, the lexical overlap between $x$ and $x_{i^{\star}}$, the agreement among the answers in $\mathcal{N}$, and the probability that the routed core assigns to the first retrieved answer token, $p_{\mathrm{core}}(a_{i^{\star},1}\mid x)$. The last feature lets the gate weigh the retrieval evidence against the core's own prediction.

\paragraph{Answer injection.}
The gate decision holds for the whole answer, with $d_t=d$ at every answer position. If the gate accepts, ExpM proposes the stored answer token by token, followed by the end-of-sequence token. At the $j$-th answer position,
\begin{equation}
p_{\mathrm{ExpM}}(y_t\mid x_{\leq t},\Delta)=\mathbf{1}\big[y_t=a_{i^{\star},j}\big],
\label{eq:mom_expm_distribution}
\end{equation}
and Eq.~\ref{eq:mom_final_distribution} is applied with $\lambda_t=0.8$. Because $p_{\mathrm{ExpM}}$ is a point mass and $\lambda_t>0.5$, the proposed token always receives the highest probability, so greedy decoding reproduces the retrieved answer exactly; the overlay then acts as gated answer copying. If the gate rejects, $d_t=0$ throughout and the routed core generates the answer alone.

\subsection{Training}
\label{subsec:mom_learning}

MoM is trained in stages so that each component is optimized for its role.

\paragraph{Teacher construction.}
We first construct a retrieval-augmented teacher distribution
\begin{equation}
p^{*}(y_t\mid x_{\leq t})=(1-\lambda_{\mathrm{ret}})\,p_{\mathrm{Base}}(y_t\mid x_{\leq t})
+\lambda_{\mathrm{ret}}\,p_{\mathrm{ret}}(y_t\mid x_{\leq t}),
\label{eq:mom_teacher}
\end{equation}
where $p_{\mathrm{ret}}$ is an offline $k$NN retrieval distribution. Interpolating with the base model reduces the brittleness of pure retrieval supervision while exposing the neural memories to retrieval-like corrections.

\paragraph{Neural memory distillation.}
LatM and LexM are trained to match the teacher while preserving next-token accuracy. For each $m\in\{\mathrm{LatM},\mathrm{LexM}\}$, we minimize
\begin{equation}
\mathcal{L}_{\mathrm{distill}}^{m}=\mu\,\operatorname{KL}\big(p^{*}\,\|\,p_m\big)+(1-\mu)\,\operatorname{CE}\big(p_m,y_t\big).
\label{eq:mom_distillation}
\end{equation}
After distillation, Base, LatM, and LexM are frozen.

\paragraph{Router training.}
With all core distributions fixed, only the router parameters $\psi$ are trained:
\begin{equation}
\begin{aligned}
\mathcal{L}_{\mathrm{router}}
={}&\mathbb{E}_{t}\big[-\log p_{\mathrm{core}}(y_t\mid x_{\leq t})\big]
+\beta_{\mathrm{LexM}}\,\mathbb{E}_{t}\big[\pi_{\mathrm{LexM}}(t)\big]\\
&+\gamma\,\mathbb{E}_{t}\big[\|\boldsymbol{\pi}(t)-\boldsymbol{\pi}(t-1)\|_1\big]
-\lambda_{\mathrm{ent}}\,\mathbb{E}_{t}\big[H(\boldsymbol{\pi}(t))\big].
\end{aligned}
\label{eq:mom_router_loss}
\end{equation}
The likelihood term teaches the router which source improves prediction. The LexM budget term discourages unnecessary use of the heavier token-space decoder, the smoothness term stabilizes decisions across adjacent tokens, and the entropy term prevents collapse to a single source.

\paragraph{Explicit memory.}
The ExpM encoders are trained once and then frozen. During this stage, $\lambda_t$ in Eq.~\ref{eq:mom_final_distribution} is produced by a token-level gate, and the encoders and gate minimize
\begin{equation}
\mathcal{L}_{\mathrm{ExpM}}=\mathcal{L}_{\mathrm{nce}}+\mathcal{L}_{\mathrm{nll}}+\mathcal{L}_{\mathrm{loc}}.
\label{eq:mom_expm_loss}
\end{equation}
$\mathcal{L}_{\mathrm{nce}}$ is an InfoNCE loss: the query at the answer start is compared with the unmasked candidate keys through $q^{\top}k_j/\tau$ with temperature $\tau$, and the cross-entropy target is the key of the question's own fact. $\mathcal{L}_{\mathrm{nll}}$ is the mean negative log-likelihood of the gold answer tokens under the mixed distribution of Eq.~\ref{eq:mom_final_distribution}, computed on examples whose fact is in the store. $\mathcal{L}_{\mathrm{loc}}$ averages $-\log(1-\lambda_t)$ over the answer positions of examples whose fact is absent from the store and of locality examples, so the gate learns to close when memory should not be used. Writing a fact $(x,a)$ requires no gradient update:
\begin{equation}
\Delta\leftarrow\Delta\cup\big\{\big(E_k(z(x)),\ \tilde{k}(x),\ a\big)\big\}.
\label{eq:mom_expm_write}
\end{equation}

At inference time, MoM computes the three core distributions, routes them into $p_{\mathrm{core}}$, and applies the ExpM overlay of Eq.~\ref{eq:mom_final_distribution} when the gate accepts the retrieved fact.

\section{Experiments}
\label{sec:experiments}

We first report factual QA results, then attribute the gains to individual memory channels, analyze how ExpM absorbs updates, and finally measure efficiency. Appendix~\ref{app:lexm_transfer} reports the transfer of LexM across backbones.

\subsection{Experimental Setup}
\label{subsec:experiment_setup}

\paragraph{Datasets.}
For language modeling we use WikiText-103~\citep{DBLP:conf/iclr/MerityX0S17}. For factual QA we evaluate on the full evaluation splits of NQ~\citep{nq}, WebQuestions (WebQ;~\citealp{webqa}), TriviaQA~\citep{triviaqa}, TruthfulQA~\citep{truthfulqa}, and HotpotQA~\citep{DBLP:conf/emnlp/Yang0ZBCSM18}. Wikipedia-2021~\citep{wikipedia2021} is the corpus for memory training, continued pretraining, the $k$NN-LM datastore, and RAG retrieval. Appendix~\ref{app:datasets} and Appendix~\ref{app:metrics} give dataset and metric details.

\paragraph{Baselines.}
We compare MoM with the frozen base LM and with two groups of methods. The parametric group comprises full continued pretraining (CPT), LoRA~\citep{lora}, and MLP Memory~\citep{mlpmemory}. The retrieval group comprises RAG with a BGE retriever~\citep{xiao2024cpack} that prepends the top-5 Wikipedia passages, $k$NN-LM with a 50M-key datastore, and a paper-guided reproduction of RAG-RL~\citep{huang2025ragrl}. MLP Memory is reported for Mistral only. Its official release provides Mistral checkpoints but no Llama-3.1-8B checkpoint, and rebuilding its backbone-specific hidden-state datastore over billions of Wikipedia tokens exceeded our storage budget. Appendix~\ref{app:baselines} specifies all configurations.

\paragraph{Implementation.}
We use Mistral-7B-v0.3~\citep{Mistral-7B} and Llama-3.1-8B~\citep{llama3} as the main backbones and GPT2-large~\citep{gpt2} for controlled language-modeling attribution. All base LMs and trained memory experts are frozen during evaluation. For the QA benchmarks, ExpM holds question--answer pairs from the training split, and its retrieval weight $\beta$, gate, and threshold $\eta$ are selected on dev folds held out from that split. Training the full stack takes about 1,800 GPU-hours on 8$\times$H200 GPUs, a one-time cost broken down by stage in Appendix~\ref{app:cost}.

\subsection{Main QA Results}
\label{subsec:main_qa_benchmark}

Table~\ref{tab:main_qa} reports the factual QA results. MoM attains the highest average on both backbones, improving the frozen backbone by 15.6\% on Mistral-7B-v0.3 and 10.9\% on Llama-3.1-8B, and leads MLP Memory, the strongest parametric baseline, by 1.08 points on Mistral. On Mistral, RAG is stronger on TruthfulQA and HotpotQA and the RAG-RL reproduction on NQ and WebQ, but MoM never falls below its frozen backbone on any benchmark, whereas every other method except MLP Memory lowers at least one benchmark. The pattern tracks where each method intervenes. CPT and LoRA rewrite the weights behind every prediction, and their WebQ gains on both backbones come with NQ losses of up to 8.47 points. RAG conditions every answer on retrieved passages and lowers the Llama average by 8.8\%. MoM instead leaves the backbone and its input unchanged and acts only on the output distribution. MoM additionally weights each memory per token from uncertainty and disagreement features and lets ExpM answer only when the gate accepts the retrieved fact, so unhelpful memory is down-weighted or skipped.

\begin{table}[t]
\centering
\caption{Factual QA results. Small colored numbers give the absolute change relative to the corresponding frozen backbone; the Average column reports the mean over the five benchmarks and its relative change. $^\dagger$Paper-guided reproduction (Appendix~\ref{app:baselines}). MLP Memory and RAG-RL are reported for Mistral only. Metric definitions are in Appendix~\ref{app:metrics}.}
\label{tab:main_qa}
\small
\setlength{\tabcolsep}{5pt}
\renewcommand{\arraystretch}{1.12}
\begin{adjustbox}{max width=\textwidth}
\begin{tabular}{lcccccc}
\toprule
\multirow{2}{*}{Methods}
& \multicolumn{3}{c}{Open-domain QA}
& \multicolumn{1}{c}{Truthfulness}
& \multicolumn{1}{c}{Multi-hop QA}
& \multirow{2}{*}{Average} \\
\cmidrule(lr){2-4}
\cmidrule(lr){5-5}
\cmidrule(lr){6-6}
& NQ & WebQ & TriviaQA & TruthfulQA & HotpotQA & \\
\midrule
Mistral-7B-v0.3
& 20.63 & 29.28 & 57.65 & 32.09 & 20.96 & 32.12 \\
\rowcolor{groupgray}
\multicolumn{7}{c}{\textit{Non-parametric methods}} \\
+ RAG
& 22.56\pos{+1.93}
& 24.90\negg{$-$4.38}
& 54.21\negg{$-$3.44}
& \textbf{35.47}\pos{+3.38}
& \textbf{29.77}\pos{+8.81}
& 33.38\pos{(+3.9\%)} \\
+ RAG-RL$^\dagger$
& \textbf{31.08}\pos{+10.45}
& \textbf{38.12}\pos{+8.84}
& 55.47\negg{$-$2.18}
& 27.46\negg{$-$4.63}
& 20.72\negg{$-$0.24}
& 34.57\pos{(+7.6\%)} \\
\rowcolor{groupgray}
\multicolumn{7}{c}{\textit{Parametric methods}} \\
+ CPT
& 12.16\negg{$-$8.47}
& 34.06\pos{+4.78}
& 61.21\pos{+3.56}
& 29.18\negg{$-$2.91}
& 16.04\negg{$-$4.92}
& 30.53\negg{($-$5.0\%)} \\
+ LoRA
& 18.17\negg{$-$2.46}
& 34.50\pos{+5.22}
& 61.60\pos{+3.95}
& 30.91\negg{$-$1.18}
& 16.23\negg{$-$4.73}
& 32.28\pos{(+0.5\%)} \\
+ MLP Memory
& 25.20\pos{+4.57}
& 37.45\pos{+8.17}
& 60.99\pos{+3.34}
& 32.54\pos{+0.45}
& 24.14\pos{+3.18}
& 36.06\pos{(+12.3\%)} \\
\rowcolor{highlightblue}
+ MoM
& 26.15\pos{+5.52}
& 37.96\pos{+8.68}
& \textbf{63.46}\pos{+5.81}
& 35.10\pos{+3.01}
& 23.01\pos{+2.05}
& \textbf{37.14}\pos{(+15.6\%)} \\
\midrule
Llama-3.1-8B
& 26.99 & 32.59 & 65.27 & 25.56 & 15.37 & 33.16 \\
\rowcolor{groupgray}
\multicolumn{7}{c}{\textit{Non-parametric methods}} \\
+ RAG
& 20.23\negg{$-$6.76}
& 25.90\negg{$-$6.69}
& 64.91\negg{$-$0.36}
& 23.93\negg{$-$1.63}
& 16.26\pos{+0.89}
& 30.25\negg{($-$8.8\%)} \\
\rowcolor{groupgray}
\multicolumn{7}{c}{\textit{Parametric methods}} \\
+ CPT
& 20.11\negg{$-$6.88}
& 36.33\pos{+3.74}
& \textbf{65.49}\pos{+0.22}
& 26.13\pos{+0.57}
& 15.42\pos{+0.05}
& 32.70\negg{($-$1.4\%)} \\
+ LoRA
& 19.40\negg{$-$7.59}
& 34.90\pos{+2.31}
& 59.50\negg{$-$5.77}
& 24.30\negg{$-$1.26}
& 11.30\negg{$-$4.07}
& 29.88\negg{($-$9.9\%)} \\
\rowcolor{highlightblue}
+ MoM
& \textbf{35.43}\pos{+8.44}
& \textbf{37.83}\pos{+5.24}
& 65.34\pos{+0.07}
& \textbf{27.98}\pos{+2.42}
& \textbf{17.22}\pos{+1.85}
& \textbf{36.76}\pos{(+10.9\%)} \\
\bottomrule
\end{tabular}
\end{adjustbox}
\end{table}

\subsection{Component Analysis}
\label{subsec:ablation_studies}

\begin{figure}[t]
\centering
\includegraphics[width=0.5\textwidth]{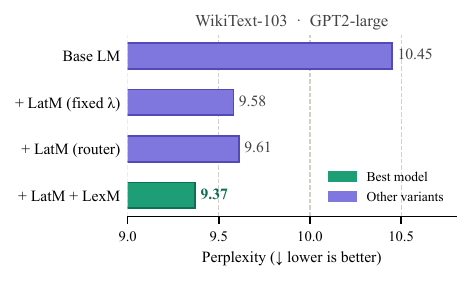}
\caption{WikiText-103 perplexity with GPT2-large and ExpM disabled (lower is better).}
\label{fig:lm_ppl}
\end{figure}

\paragraph{Language modeling channel attribution.}
Figure~\ref{fig:lm_ppl} isolates the neural memories on WikiText-103 with GPT2-large and ExpM disabled. LatM with a fixed mixing weight lowers perplexity from 10.45 to 9.58. A token-wise router over Base LM and LatM reaches 9.61, recovering most of this gain without a globally tuned coefficient. Adding LexM lowers perplexity further to 9.37, so the token-space memory contributes information that LatM does not provide. The result shows that the routed core assigns more probability to the observed next token, and with the backbone frozen, this gain reflects knowledge that the neural memories add without changing the backbone weights. Appendix~\ref{app:gates} reports how the router distributes weight across token types on each backbone.

\begin{figure}[t]
\centering
\begin{minipage}[t]{0.48\textwidth}
\centering
\includegraphics[width=\linewidth]{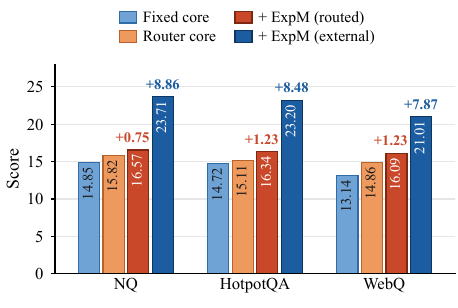}
\caption{Explicit memory ablation study.}
\label{fig:expm_route}
\end{minipage}\hfill
\begin{minipage}[t]{0.48\textwidth}
\centering
\includegraphics[width=\linewidth]{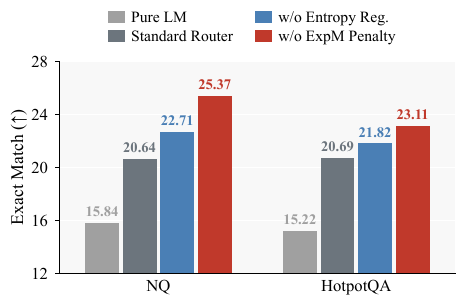}
\caption{EM of router variants that route ExpM as a fourth source with the store populated at inference.}
\label{fig:router_ablation}
\end{minipage}
\end{figure}

\paragraph{Routed versus external explicit memory.}
Figure~\ref{fig:expm_route} tests the central design choice of MoM by exposing the same populated store in two ways. As an external overlay, ExpM raises QA score by 7.87--8.86 points on the three benchmarks. As a fourth routed source, the same store adds only 0.75--1.23 points, because the router, trained while the store was empty or sparse, assigns ExpM little mass. The store contents are identical, so the gap reflects allocation rather than memory quality.

\paragraph{Router training with routed ExpM.}
Figure~\ref{fig:router_ablation} shows that the router's own training objective limits a routed ExpM. Every variant improves over Pure LM, removing either regularizer helps further, and removing the ExpM usage penalty helps most. Both terms oppose the allocation ExpM needs. A stored answer changes a greedy prediction only when its weight lets the stored token win, as the overlay ensures with $\lambda_t=0.8$. The entropy regularizer, which keeps the router from collapsing onto one source, penalizes such peaked allocations, and the usage penalty lowers ExpM mass directly. Learned while the store is empty or sparse, the resulting low ExpM weight persists once the store is populated. Using a stored fact is an all-or-nothing decision per question, which suits an answer-level gate better than a token-level softmax trained before the fact is written.

\subsection{Explicit Memory Updates}
\label{subsec:expm_analysis}

We test whether written facts take effect immediately and whether writing them disturbs answers the model already gets right. For each backbone, we draw two disjoint sets of 200 questions from the NQ-open training split, excluding the dev folds and all questions used to train or calibrate ExpM: $D_{\text{new}}$, which both the base LM and the routed core answer incorrectly, and $D_{\text{old}}$, which both answer correctly. We write the $D_{\text{new}}$ facts, one answer each, into an empty store and evaluate both sets with the QA configuration of Table~\ref{tab:main_qa}. Since $D_{\text{new}}$ is queried with the questions used for writing, its score measures how reliably a written fact is read back, not generalization to rephrased questions.

\begin{figure}[t]
\centering
\begin{minipage}[t]{0.48\textwidth}
\centering
\includegraphics[width=\linewidth]{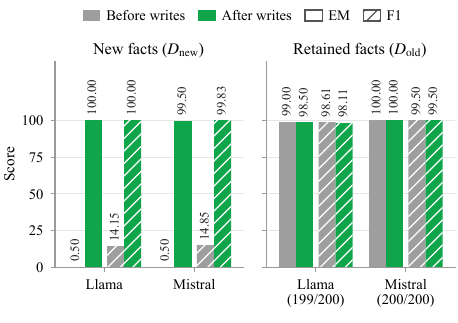}
\caption{EM and F1 before and after writing the 200 $D_{\text{new}}$ facts into an initially empty ExpM store. $D_{\text{new}}$ is queried with the questions used for writing. Parentheses give the number of $D_{\text{old}}$ questions whose decoded answer is unchanged.}
\label{fig:update}
\end{minipage}\hfill
\begin{minipage}[t]{0.48\textwidth}
\centering
\includegraphics[width=\linewidth]{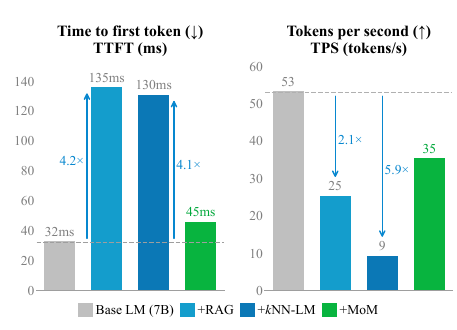}
\caption{Inference efficiency with a 7B backbone, measured by time to first token (TTFT, lower is better) and tokens per second (TPS, higher is better).}
\label{fig:latency_tps}
\end{minipage}
\end{figure}

Figure~\ref{fig:update} shows that the writes take effect immediately and leave $D_{\text{old}}$ almost untouched. EM on $D_{\text{new}}$ rises from 0.50 on both backbones to 100.00 on Llama and 99.50 on Mistral. Since no $D_{\text{old}}$ fact is written, a $D_{\text{old}}$ answer can change only if the gate accepts an entry retrieved for an unrelated question. This happens once in 400 queries: on Llama, the release year of \textit{Call Me by Your Name} changes from the correct 2017 to ``May 11, 2018,'' lowering EM from 99.00 to 98.50.

\subsection{Efficiency}
\label{subsec:efficiency}

Figure~\ref{fig:latency_tps} compares time to first token (TTFT) and decoding throughput (TPS). RAG and $k$NN-LM raise TTFT from 32\,ms to 135 and 130\,ms and lower throughput from 53 to 25 and 9 tokens/s. Both put a search over an external index on the critical path, RAG also prefills the retrieved passages, and $k$NN-LM searches its datastore at every decoding step. MoM with the full router reaches 45\,ms and 35 tokens/s, a 2.9--3.0$\times$ lower TTFT than the two retrieval methods. Compared with the frozen base model, this is a 1.4$\times$ higher TTFT and 66\% of its decoding throughput. Appendix~\ref{app:cost} reports memory, storage, and training costs.

\section{Conclusion}
\label{sec:conclusion}

We presented Mixture-of-Memories (MoM), a memory-augmented language modeling framework that routes stable neural memories token by token and handles factual updates through a writable explicit memory outside the router. MoM lowers WikiText-103 perplexity and obtains the best average factual QA score on Mistral-7B-v0.3 and Llama-3.1-8B, with a lower time to first token than retrieval-augmented baselines. The same populated store helps far more as an external overlay than as a routed source, and written facts take effect immediately while previously correct answers are almost all preserved. Future work can extend MoM to other knowledge-intensive NLP tasks and to settings where knowledge changes continuously.

\section*{Limitations}

The proposed method has two main limitations. First, the explicit memory selects a single fact for each question from the whole store: as the store grows, more entries compete for each query, which raises the chance that a wrong fact is retrieved and accepted, making the current design more suitable for targeted factual updates than very large memory stores. Second, although our results are strong on QA benchmarks, the present evaluation is still concentrated on knowledge-intensive question answering. Further validation on a broader set of NLP tasks is needed to assess the generality of the approach.





%% file: sections/appendix.tex
\section{Experimental Details}
\label{app:experiment_setup}

\subsection{Datasets and Splits}
\label{app:datasets}

WikiText-103 is used for GPT2-large language modeling. Wikipedia-2021 is used for memory training, CPT, the $k$NN-LM datastore, and the RAG corpus. The QA evaluation uses the full evaluation splits of Natural Questions (3,610 questions), HotpotQA (7,405), WebQuestions (2,032), TriviaQA (17,944), and TruthfulQA. For the update experiments, $D_{\mathrm{new}}$ and $D_{\mathrm{old}}$ each contain 200 questions from the NQ-open training split, selected separately for each backbone (Section~\ref{subsec:expm_analysis}). The facts of $D_{\mathrm{new}}$ are written into an initially empty store, and $D_{\mathrm{old}}$ is never written. We report score after all 200 writes.

\subsection{Baselines}
\label{app:baselines}

\paragraph{Parametric baselines.}
CPT updates all parameters of the backbone by continued pretraining with the language-modeling objective. LoRA~\citep{lora} keeps the backbone frozen and learns low-rank updates to its weight matrices. MLP Memory~\citep{mlpmemory} is a pretrained MLP that imitates the next-token distribution of a nearest-neighbor retriever, and its output is interpolated with the distribution of the frozen backbone.

\paragraph{Retrieval baselines.}
RAG retrieves the top-5 Wikipedia passages for each question with a BGE retriever~\citep{xiao2024cpack} and prepends them to the backbone's prompt. $k$NN-LM~\citep{DBLP:conf/iclr/KhandelwalLJZL20} interpolates the backbone's next-token distribution with the next tokens of the nearest stored contexts, retrieved from a 50M-key datastore stored with product quantization.

\paragraph{Post-trained RAG reproduction.}
RAG-RL~\citep{huang2025ragrl} trains the reader with reinforcement learning and curriculum learning. We implemented a paper-guided reproduction with a \texttt{bge-base-en-v1.5} retriever, exact inner-product FAISS search over 200,000 uniformly sampled Wikipedia-2021 chunks, question-only queries, top-5 retrieval, greedy bf16 decoding, and seed 42, with Mistral-7B-v0.3 as the reader backbone.

\subsection{Metrics}
\label{app:metrics}

For open-domain QA, EM is computed after answer normalization and measures whether the generated answer exactly matches any accepted answer alias. F1 is the token-level overlap after the same normalization. We report F1 for NQ. For TruthfulQA, we use the official multiple-choice suite: MC1 measures whether the single best answer is preferred, MC2 measures the total probability assigned to true answers relative to false answers, and MC3 measures the fraction of true answers ranked above false answers. We report their average when a single TruthfulQA score is needed.

\subsection{ExpM Configuration}
\label{app:implementation}

The Budget Router outputs weights over Base, LatM, and LexM only, and the routed core is trained separately from ExpM. The ExpM encoders read the frozen backbone feature of Eq.~\ref{eq:mom_expm_feature} with a 32-token averaging window. For the QA benchmarks, ExpM retrieves the top-scoring entry over the whole store with the score of Eq.~\ref{eq:mom_expm_score}, computes the agreement feature from the top-$k$ entries, and injects the retrieved answer with $\lambda=0.8$ when the gate accepts it. On the dev folds, $\beta$ is chosen from $\{0, 0.25, 0.5, 0.75, 1\}$ and $k$ from $\{4, 8, 16\}$, so the $\beta$ used at test time can differ across benchmarks and backbones. TriviaQA answers have many valid surface forms, so each TriviaQA training question is written with up to ten answer aliases; evaluation still uses the official normalized scorer, which accepts any annotated alias. The encoders are not retrained for the QA experiments. Only the answer-level gate is fitted.

\subsection{Hyperparameter Selection}
\label{app:hyperparameters}

In the QA experiments, the gate threshold is $\eta=0.35$ for NQ, WebQ, and HotpotQA on both backbones, and $0.5$ (Mistral) and $0.8$ (Llama) for TriviaQA. Ablations use $\lambda=0.4$ unless they sweep the value. We report full test-set results for the main tables.

\subsection{Training Cost and Deployment Overhead}
\label{app:cost}

Training takes about 1,800 H200 GPU-hours once: about 400 for the $k$NN teacher targets, 1,100 for LatM distillation, 150 for LexM, 50 for the router, and 100 for preprocessing and pipeline overhead. The stages run sequentially with the base LM frozen, and afterwards new facts enter only through ExpM writes.

With a 7B backbone, LatM (1.0B parameters, 2.0\,GB in bf16), LexM (0.77B, 1.55\,GB), and the 580-parameter router raise resident weights from 14.0\,GB to 17.6\,GB. The ExpM store has no trainable parameters, occupies about 15\,MB, and grows with the number of written facts, whereas our product-quantized $k$NN-LM datastore occupies 3.2\,GB. Both neural memories are MLPs that add about $6\cdot n_{\text{layer}}\cdot d_{\text{model}}\cdot d_{\text{ff}}$ FLOPs per token, and ExpM retrieves once per question rather than per token.

\section{Supplementary Experiments}
\label{app:more_exp}

\subsection{Cross-Backbone Transfer of Lexical Memory}
\label{app:lexm_transfer}

\begin{table}[t]
\centering
\caption{LexM attached to Llama-3.1-8B. ``Mistral'' rows transplant a Mistral-trained LexM through a token-string vocabulary mapping without retraining; ``Llama'' rows use a LexM trained on Llama. $\lambda$ is a fixed mixing weight.}
\label{tab:lexm_transfer}
\small
\setlength{\tabcolsep}{4pt}
\begin{adjustbox}{max width=\textwidth}
\begin{tabular}{lcccccc}
\toprule
\multirow{2}{*}{LexM on Llama} & \multicolumn{2}{c}{NQ} & \multicolumn{2}{c}{HotpotQA} & \multicolumn{2}{c}{TriviaQA} \\
\cmidrule(lr){2-3}\cmidrule(lr){4-5}\cmidrule(lr){6-7}
& EM & F1 & EM & F1 & EM & F1 \\
\midrule
None (Llama base) & 17.15 & 26.99 & 15.37 & 25.09 & 65.27 & 71.61 \\
\midrule
Mistral, $\lambda=0.10$ & 16.95 & 26.99 & 15.34 & 25.16 & 65.21 & 71.61 \\
Mistral, small router & 16.95 & 26.99 & 15.37 & 25.17 & 65.24 & \textbf{71.64} \\
\midrule
Llama, $\lambda=0.10$ & 18.34 & 28.17 & \textbf{15.67} & \textbf{25.35} & \textbf{65.30} & 71.61 \\
Llama, $\lambda=0.25$ & \textbf{19.47} & \textbf{29.17} & 15.48 & 24.99 & 64.92 & 71.09 \\
\bottomrule
\end{tabular}
\end{adjustbox}
\end{table}

LexM consumes token IDs, so it can be moved to another backbone. We transplant a Mistral-trained LexM into Llama-3.1-8B through a token-string mapping between the two vocabularies, which changes 8.27\% of the target embedding rows and requires no retraining. Table~\ref{tab:lexm_transfer} shows that the transplanted LexM keeps EM within 0.20 points of the Llama backbone on all three benchmarks, so it can be attached without disrupting the base model but adds no knowledge in this setting. In contrast, a LexM trained on Llama itself raises NQ EM from 17.15 to 19.47 with a fixed weight of 0.25, while changing TriviaQA EM by at most 0.35 points.

\subsection{Router Gate Interpretability}
\label{app:gates}

To test whether the router learns non-trivial token-wise allocation, we group generated tokens into semantic, numeric, and syntactic categories, represented by ENTITY, NUMBER, and PUNCT tokens, and compute the average gate weight of each memory channel. The allocation depends on both the token type and the backbone. On Mistral, punctuation tokens are routed mostly to the base LM (gate weight 0.908 on PUNCT), while NUMBER tokens receive a larger LatM weight of 0.286, so LatM is used selectively on factual or numeric positions. On Llama, LexM dominates, with gate weights of 0.614, 0.698, and 0.749 on ENTITY, NUMBER, and PUNCT, while LatM receives only 0.7--1.2\% of the mass. The router therefore adapts its memory usage to token semantics and to the quality of each channel on a given backbone rather than following a static interpolation rule.

\subsection{Case Study}
\label{app:case}
\begin{figure}[t]
\centering
\includegraphics[width=\textwidth]{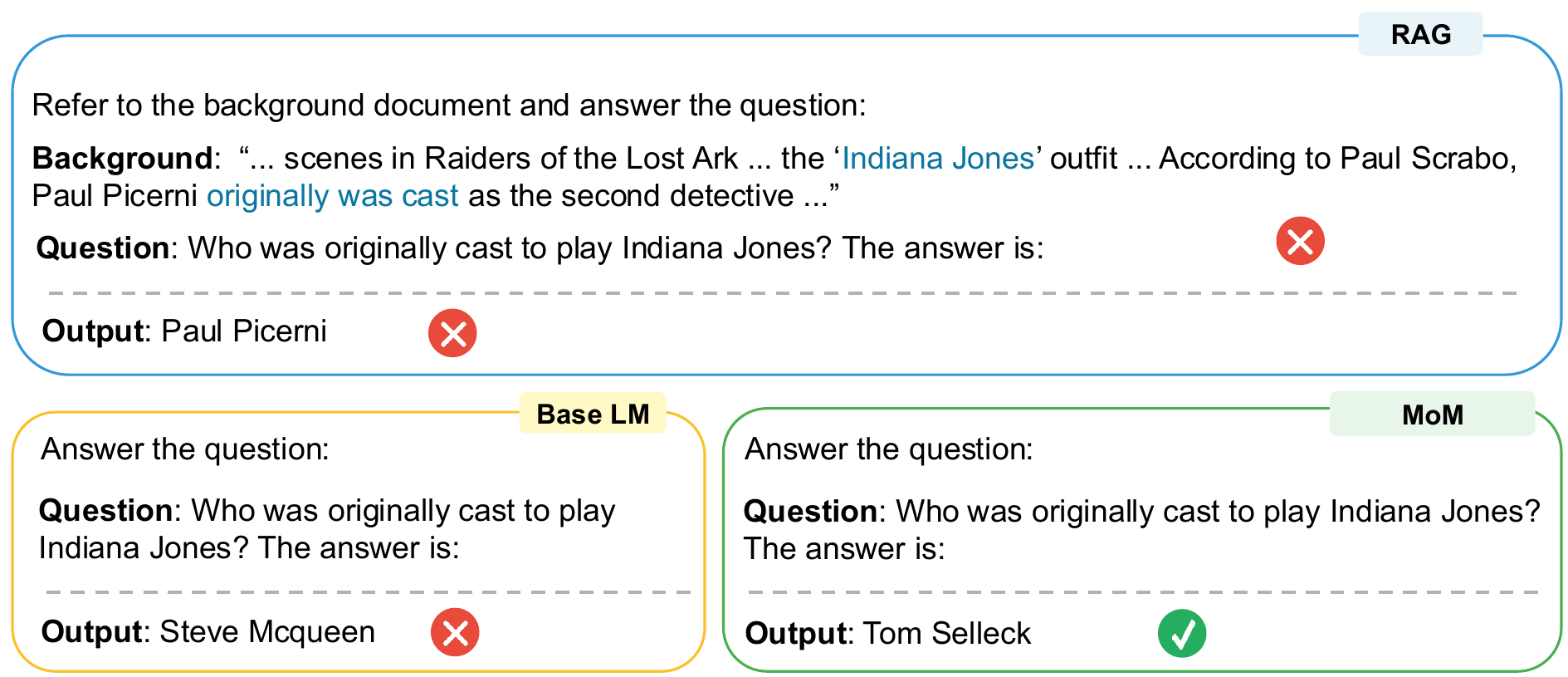}
\caption{Model outputs on a factual question. RAG retrieves superficially relevant content but extracts the wrong entity, while MoM returns the correct answer.}
\label{fig:case}
\end{figure}

Figure~\ref{fig:case} shows a case study. For the question of who was originally cast as Indiana Jones, RAG retrieves passages that contain both ``Indiana Jones'' and the phrase ``originally was cast.'' The retrieved evidence, however, states that Paul Picerni was originally cast in a different role. RAG copies this entity and answers ``Paul Picerni,'' the base LM answers ``Steve McQueen,'' and CPT, LoRA, and MLP Memory also return plausible but incorrect actors. MoM returns the gold answer, ``Tom Selleck,'' showing that ExpM can recover a written fact even when lexical retrieval is misled by entity and phrase overlap.

%% file: main.bib
@article{selfparam,
  title={Self-Updatable Large Language Models by Integrating Context into Model Parameters},
  author={Wang, Yu and Liu, Xinshuang and Chen, Xiusi and O'Brien, Sean and Wu, Junda and McAuley, Julian},
  journal={arXiv preprint arXiv:2410.00487},
  year={2024}
}

@article{memoryllm,
  title={{MemoryLLM}: Towards Self-Updatable Large Language Models},
  author={Wang, Yu and Gao, Yifan and Chen, Xiusi and Jiang, Haoming and Li, Shiyang and Yang, Jingfeng and Yin, Qingyu and Li, Zheng and Li, Xian and Yin, Bing and others},
  journal={arXiv preprint arXiv:2402.04624},
  year={2024}
}

@article{mplus,
  title={{M+}: Extending {MemoryLLM} with Scalable Long-Term Memory},
  author={Wang, Yu and Krotov, Dmitry and Hu, Yuanzhe and Gao, Yifan and Zhou, Wangchunshu and McAuley, Julian and Gutfreund, Dan and Feris, Rogerio and He, Zexue},
  journal={arXiv preprint arXiv:2502.00592},
  year={2025}
}

@article{lewis2020rag,
  title={Retrieval-Augmented Generation for Knowledge-Intensive {NLP} Tasks},
  author={Lewis, Patrick and Perez, Ethan and Piktus, Aleksandra and Petroni, Fabio and Karpukhin, Vladimir and Goyal, Naman and K{\"u}ttler, Heinrich and Lewis, Mike and Yih, Wen-tau and Rockt{\"a}schel, Tim and others},
  journal={Advances in Neural Information Processing Systems},
  volume={33},
  pages={9459--9474},
  year={2020}
}

@inproceedings{borgeaud2022retro,
  title={Improving Language Models by Retrieving from Trillions of Tokens},
  author={Borgeaud, Sebastian and Mensch, Arthur and Hoffmann, Jordan and Cai, Trevor and Rutherford, Eliza and Millican, Katie and Van Den Driessche, George Bm and Lespiau, Jean-Baptiste and Damoc, Bogdan and Clark, Aidan and others},
  booktitle={International Conference on Machine Learning},
  pages={2206--2240},
  year={2022},
  organization={PMLR}
}

@inproceedings{DBLP:conf/iclr/KhandelwalLJZL20,
  author       = {Urvashi Khandelwal and
                  Omer Levy and
                  Dan Jurafsky and
                  Luke Zettlemoyer and
                  Mike Lewis},
  title        = {Generalization through Memorization: Nearest Neighbor Language Models},
  booktitle    = {8th International Conference on Learning Representations, {ICLR} 2020,
                  Addis Ababa, Ethiopia, April 26-30, 2020},
  publisher    = {OpenReview.net},
  year         = {2020}
}

@article{mlpmemory,
  title={{MLP} Memory: A Retriever-Pretrained Memory for Large Language Models},
  author={Wei, Rubin and Cao, Jiaqi and Wang, Jiarui and Kai, Jushi and Guo, Qipeng and Zhou, Bowen and Lin, Zhouhan},
  journal={arXiv preprint arXiv:2508.01832},
  year={2025}
}

@article{cao2026memory,
  title={Memory decoder: A pretrained, plug-and-play memory for large language models},
  author={Cao, Jiaqi and Wang, Jiarui and Wei, Rubin and Guo, Qipeng and Chen, Kai and Zhou, Bowen and Lin, Zhouhan},
  journal={Advances in Neural Information Processing Systems},
  volume={38},
  pages={115487--115510},
  year={2026}
}

@inproceedings{yao2025seakr,
  title={Seakr: Self-aware knowledge retrieval for adaptive retrieval augmented generation},
  author={Yao, Zijun and Qi, Weijian and Pan, Liangming and Cao, Shulin and Hu, Linmei and Weichuan, Liu and Hou, Lei and Li, Juanzi},
  booktitle={Proceedings of the 63rd Annual Meeting of the Association for Computational Linguistics (Volume 1: Long Papers)},
  pages={27022--27043},
  year={2025}
}

@inproceedings{DBLP:conf/iclr/ShazeerMMDLHD17,
  author       = {Noam Shazeer and
                  Azalia Mirhoseini and
                  Krzysztof Maziarz and
                  Andy Davis and
                  Quoc V. Le and
                  Geoffrey E. Hinton and
                  Jeff Dean},
  title        = {Outrageously Large Neural Networks: The Sparsely-Gated Mixture-of-Experts
                  Layer},
  booktitle    = {5th International Conference on Learning Representations, {ICLR} 2017,
                  Toulon, France, April 24-26, 2017, Conference Track Proceedings},
  publisher    = {OpenReview.net},
  year         = {2017}
}

@article{fedus2022switch,
  title={Switch Transformers: Scaling to Trillion Parameter Models with Simple and Efficient Sparsity},
  author={Fedus, William and Zoph, Barret and Shazeer, Noam},
  journal={Journal of Machine Learning Research},
  volume={23},
  number={120},
  pages={1--39},
  year={2022}
}

@inproceedings{du2022glam,
  title={{GLaM}: Efficient Scaling of Language Models with Mixture-of-Experts},
  author={Du, Nan and Huang, Yanping and Dai, Andrew M and Tong, Simon and Lepikhin, Dmitry and Xu, Yuanzhong and Krikun, Maxim and Zhou, Yanqi and Yu, Adams Wei and Firat, Orhan and others},
  booktitle={International Conference on Machine Learning},
  pages={5547--5569},
  year={2022},
  organization={PMLR}
}

@inproceedings{DBLP:conf/iclr/MitchellLBFM22,
  author       = {Eric Mitchell and
                  Charles Lin and
                  Antoine Bosselut and
                  Chelsea Finn and
                  Christopher D. Manning},
  title        = {Fast Model Editing at Scale},
  booktitle    = {The Tenth International Conference on Learning Representations, {ICLR}
                  2022, Virtual Event, April 25-29, 2022},
  publisher    = {OpenReview.net},
  year         = {2022}
}

@article{meng2022rome,
  title={Locating and Editing Factual Associations in {GPT}},
  author={Meng, Kevin and Bau, David and Andonian, Alex and Belinkov, Yonatan},
  journal={Advances in Neural Information Processing Systems},
  volume={35},
  pages={17359--17372},
  year={2022}
}

@inproceedings{DBLP:conf/iclr/MengSABB23,
  author       = {Kevin Meng and
                  Arnab Sen Sharma and
                  Alex J. Andonian and
                  Yonatan Belinkov and
                  David Bau},
  title        = {Mass-Editing Memory in a Transformer},
  booktitle    = {The Eleventh International Conference on Learning Representations,
                  {ICLR} 2023, Kigali, Rwanda, May 1-5, 2023},
  publisher    = {OpenReview.net},
  year         = {2023}
}

@inproceedings{mitchell2022serac,
  title={Memory-Based Model Editing at Scale},
  author={Mitchell, Eric and Lin, Charles and Bosselut, Antoine and Manning, Christopher D and Finn, Chelsea},
  booktitle={International Conference on Machine Learning},
  pages={15817--15831},
  year={2022},
  organization={PMLR}
}

@article{hartvigsen2023grace,
  title={Aging with {GRACE}: Lifelong Model Editing with Discrete Key-Value Adaptors},
  author={Hartvigsen, Tom and Sankaranarayanan, Swami and Palangi, Hamid and Kim, Yoon and Ghassemi, Marzyeh},
  journal={Advances in Neural Information Processing Systems},
  volume={36},
  pages={47934--47959},
  year={2023}
}

@article{wang2024wise,
  title={{WISE}: Rethinking the Knowledge Memory for Lifelong Model Editing of Large Language Models},
  author={Wang, Peng and Li, Zexi and Zhang, Ningyu and Xu, Ziwen and Yao, Yunzhi and Jiang, Yong and Xie, Pengjun and Huang, Fei and Chen, Huajun},
  journal={Advances in Neural Information Processing Systems},
  volume={37},
  pages={53764--53797},
  year={2024}
}

@inproceedings{lora,
  title={{LoRA}: Low-Rank Adaptation of Large Language Models},
  author={Hu, Edward J and Shen, Yelong and Wallis, Phillip and Allen-Zhu, Zeyuan and Li, Yuanzhi and Wang, Shean and Wang, Lu and Chen, Weizhu},
  booktitle={International Conference on Learning Representations},
  year={2022}
}

@inproceedings{jiao2025hirag,
  title={{HIRAG}: Hierarchical-Thought Instruction-Tuning Retrieval-Augmented Generation},
  author={Jiao, Yihan and Tan, Zhehao and Yang, Dan and Sun, Duolin and Feng, Jie and Shen, Yue and Wang, Jian and Wei, Peng},
  booktitle={Findings of the Association for Computational Linguistics: EMNLP 2025},
  pages={5111--5130},
  year={2025},
  url={https://aclanthology.org/2025.findings-emnlp.274/}
}

@article{huang2025ragrl,
  title={{RAG-RL}: Advancing Retrieval-Augmented Generation via {RL} and Curriculum Learning},
  author={Huang, Jerry and Madala, Siddarth and Sidhu, Risham and Niu, Cheng and Peng, Hao and Hockenmaier, Julia and Zhang, Tong},
  journal={arXiv preprint arXiv:2503.12759},
  year={2025},
  url={https://arxiv.org/abs/2503.12759}
}

@inproceedings{xiao2024cpack,
  title={{C-Pack}: Packed Resources for General {C}hinese Embeddings},
  author={Xiao, Shitao and Liu, Zheng and Zhang, Peitian and Muennighoff, Niklas and Lian, Defu and Nie, Jian-Yun},
  booktitle={Proceedings of the 47th International ACM SIGIR Conference on Research and Development in Information Retrieval},
  year={2024},
  url={https://arxiv.org/abs/2309.07597}
}

@article{nq,
  title={Natural Questions: A Benchmark for Question Answering Research},
  author={Kwiatkowski, Tom and Palomaki, Jennimaria and Redfield, Olivia and Collins, Michael and Parikh, Ankur and Alberti, Chris and Epstein, Danielle and Polosukhin, Illia and Devlin, Jacob and Lee, Kenton and others},
  journal={Transactions of the Association for Computational Linguistics},
  volume={7},
  pages={453--466},
  year={2019}
}

@inproceedings{triviaqa,
  title={{TriviaQA}: A Large Scale Distantly Supervised Challenge Dataset for Reading Comprehension},
  author={Joshi, Mandar and Choi, Eunsol and Weld, Daniel S and Zettlemoyer, Luke},
  booktitle={Proceedings of the 55th Annual Meeting of the Association for Computational Linguistics (Volume 1: Long Papers)},
  pages={1601--1611},
  year={2017}
}

@inproceedings{webqa,
  title={Semantic Parsing on {F}reebase from Question-Answer Pairs},
  author={Berant, Jonathan and Chou, Andrew and Frostig, Roy and Liang, Percy},
  booktitle={Proceedings of the 2013 Conference on Empirical Methods in Natural Language Processing},
  pages={1533--1544},
  year={2013}
}

@inproceedings{DBLP:conf/emnlp/Yang0ZBCSM18,
  author       = {Zhilin Yang and
                  Peng Qi and
                  Saizheng Zhang and
                  Yoshua Bengio and
                  William W. Cohen and
                  Ruslan Salakhutdinov and
                  Christopher D. Manning},
  editor       = {Ellen Riloff and
                  David Chiang and
                  Julia Hockenmaier and
                  Jun'ichi Tsujii},
  title        = {HotpotQA: {A} Dataset for Diverse, Explainable Multi-hop Question
                  Answering},
  booktitle    = {Proceedings of the 2018 Conference on Empirical Methods in Natural
                  Language Processing, Brussels, Belgium, October 31 - November 4, 2018},
  pages        = {2369--2380},
  publisher    = {Association for Computational Linguistics},
  year         = {2018},
}

@inproceedings{truthfulqa,
  title={{TruthfulQA}: Measuring How Models Mimic Human Falsehoods},
  author={Lin, Stephanie and Hilton, Jacob and Evans, Owain},
  booktitle={Proceedings of the 60th Annual Meeting of the Association for Computational Linguistics (Volume 1: Long Papers)},
  pages={3214--3252},
  year={2022}
}

@article{wikipedia2021,
  title={Atlas: Few-Shot Learning with Retrieval Augmented Language Models},
  author={Izacard, Gautier and Lewis, Patrick and Lomeli, Maria and Hosseini, Lucas and Petroni, Fabio and Schick, Timo and Dwivedi-Yu, Jane and Joulin, Armand and Riedel, Sebastian and Grave, Edouard},
  journal={Journal of Machine Learning Research},
  volume={24},
  number={251},
  pages={1--43},
  year={2023}
}

@inproceedings{DBLP:conf/iclr/MerityX0S17,
  author       = {Stephen Merity and
                  Caiming Xiong and
                  James Bradbury and
                  Richard Socher},
  title        = {Pointer Sentinel Mixture Models},
  booktitle    = {5th International Conference on Learning Representations, {ICLR} 2017,
                  Toulon, France, April 24-26, 2017, Conference Track Proceedings},
  publisher    = {OpenReview.net},
  year         = {2017}
}

@misc{Mistral-7B,
  title={Mistral {7B}},
  author={Albert Q. Jiang and Alexandre Sablayrolles and Arthur Mensch and Chris Bamford and Devendra Singh Chaplot and Diego de las Casas and Florian Bressand and Gianna Lengyel and Guillaume Lample and Lucile Saulnier and L{\'e}lio Renard Lavaud and Marie-Anne Lachaux and Pierre Stock and Teven Le Scao and Thibaut Lavril and Thomas Wang and Timoth{\'e}e Lacroix and William El Sayed},
  year={2023},
  eprint={2310.06825},
  archivePrefix={arXiv},
  primaryClass={cs.CL},
  url={https://arxiv.org/abs/2310.06825}
}

@article{llama3,
  title={The {Llama} 3 Herd of Models},
  author={Grattafiori, Aaron and Dubey, Abhimanyu and Jauhri, Abhinav and Pandey, Abhinav and Kadian, Abhishek and Al-Dahle, Ahmad and Letman, Aiesha and Mathur, Akhil and Schelten, Alan and Vaughan, Alex and others},
  journal={arXiv preprint arXiv:2407.21783},
  year={2024}
}

@article{gpt2,
  title={Language Models are Unsupervised Multitask Learners},
  author={Radford, Alec and Wu, Jeffrey and Child, Rewon and Luan, David and Amodei, Dario and Sutskever, Ilya and others},
  journal={OpenAI blog},
  volume={1},
  number={8},
  pages={9},
  year={2019}
}

@inproceedings{allen2025physics,
  title={Physics of language models: Part 3.2, knowledge manipulation},
  author={Allen-Zhu, Zeyuan and Li, Yuanzhi},
  booktitle={International Conference on Learning Representations},
  volume={2025},
  pages={85824--85832},
  year={2025}
}

@article{chang2024acquire,
  title={How do large language models acquire factual knowledge during pretraining?},
  author={Chang, Hoyeon and Park, Jinho and Ye, Seonghyeon and Yang, Sohee and Seo, Youngkyung and Chang, Du-Seong and Seo, Minjoon},
  journal={Advances in neural information processing systems},
  volume={37},
  pages={60626--60668},
  year={2024}
}

@inproceedings{gekhman2024finetuning,
  title={Does fine-tuning LLMs on new knowledge encourage hallucinations?},
  author={Gekhman, Zorik and Yona, Gal and Aharoni, Roee and Eyal, Matan and Feder, Amir and Reichart, Roi and Herzig, Jonathan},
  booktitle={Proceedings of the 2024 conference on empirical methods in natural language processing},
  pages={7765--7784},
  year={2024}
}

@inproceedings{ovadia2024finetuning,
  title={Fine-tuning or retrieval? comparing knowledge injection in llms},
  author={Ovadia, Oded and Brief, Menachem and Mishaeli, Moshik and Elisha, Oren},
  booktitle={Proceedings of the 2024 conference on empirical methods in natural language processing},
  pages={237--250},
  year={2024}
}
